# CLASSIFICATION OF SPHERICAL OBJECTS BASED ON THE FORM FUNCTION OF ACOUSTIC ECHOES

Mariia Dmitrieva[a],   Keith E. Brown[b], Gary J. Heald[c], David M. Lane[d]

 Heriot-Watt University, School of Engineering and Physical Sciences, Earl Mounbatten Building, Edinburgh EH11 4AS United Kingdom
[a] m.dmitrieva@hw.ac.uk
[b] k.e.brown@hw.ac.uk
[c] gjheald@dstl.gov.uk
[d] d.m.lane@hw.ac.uk

**Abstract:** *One way to recognise an object is to study how the echo has been shaped during the interaction with the target. Using wideband sonar allows the study of the energy distribution for a large range of frequencies. The frequency distribution contains information about an object, including its inner structure. This information is a key for automatic recognition.*

*The scattering by a target can be quantitatively described by its Form Function. The Form Function can be calculated based on the data of the initial pulse, reflected pulse and parameters of a medium where the pulse is propagating.*

*In this work spherical objects are classified based on their filler material – water or air. We limit the study to spherical 2 layered targets immersed in water.*

*The Form Function is used as a descriptor and fed into a Neural Network classifier, Multilayer Perceptron (MLP). The performance of the classifier is compared with Support Vector Machine (SVM) and the Form Function descriptor is examined in contrast to the Time and Frequency Representation of the echo.*

## 1. INTRODUCTION

The majority of sonars use narrow bandwidth pulses and measure timing of the returned ping. Wideband sonar brings the possibility to study the broadband echo structure. In this work we use wideband pulses to compute Form Function of an object.

The Form Function describes scattering from a target. It is a signature of an object which depends on the object's properties including shape, size, material of the object's shell and material of the object's filler. Study of the scattering has a long history.

Form Function analytical solution for simple shapes was presented by Faran [2]. Hickling [3] introduced a solution for a solid elastic sphere. Goodman [4] calculates reflection for a spherical shell with a fluid filling. Doolittle's work [5] shows a solution for a cylinder shell. Chinnery [6] describes acoustic scattering from a cube shape. Pailhas [7] presented a solution for a multi layered spheres.

In this work we use a two layer sphere as a target. The spherical target is a simple geometrical shape with a spherical symmetry. This shape provides independence from the view angle.

The Form Function is handled as a descriptor of the object. It has distinguishing peaks and notches. Positions of the peaks and notches are related to the object's properties. The targets are classified based on the Form Function with the MLP Neural Network Classifier. The approach provides object classification based on their filler material.

There are number of approaches which provide insight for object classification based on the target scattering [8], [9], [10], and [11]. The novelty of this method is the combination of the Form Function and a Neural Network classifier for the filler material based classification.

The paper compares results of the MLP classifier with the SVM classifier for the same dataset and object descriptors. Using the Form Function as a primary object descriptor, we compare it with an echo representation in Time and Frequency Domains.

## 2. METHODOLOGY

The object classification approach can be presented in three steps: scattering segmentation, calculation of the Form Function and classification, Figure 1.

### 2.1 ECHO SEGMENTATION

The recordings are made in a 3 x 4 x 2 m water tank. The target is fixed in the water in a range from 1.5 to 3 metres away from the sonar. The objects are located in a far-field of the sonar's transducer.

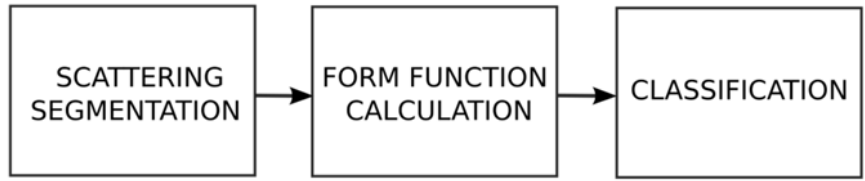

*Fig.1: Classification scheme.*

Due to the size of the water tank the recorded data contains reflection from the object, walls, bottom of the tank and other surfaces, Figure 2.

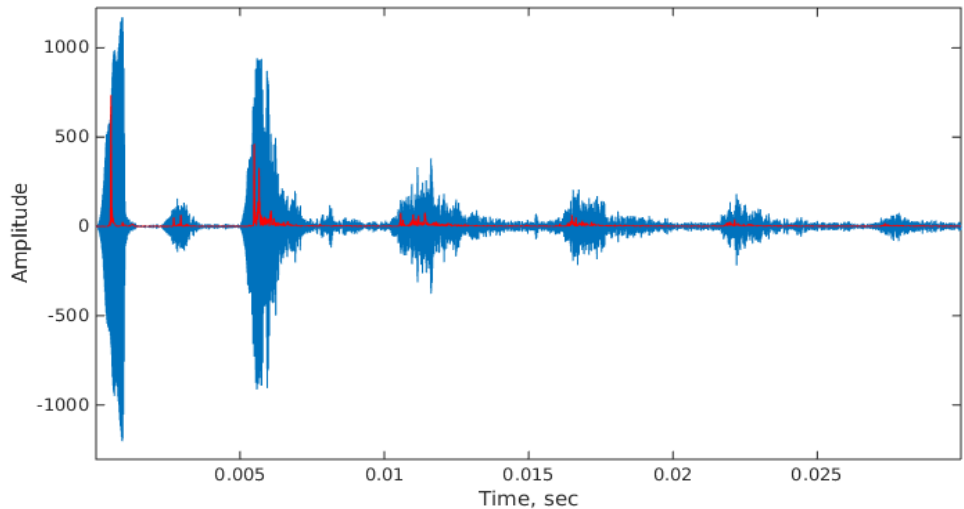

*Fig.1: Recording of a response (blue) and matched filtering output (red) of the first 30 ms of a recording.*

The echo from the object is segmented using a matched filter. Based on the experimental set-up only one object is located in the range from 1.5 to 3 metres, corresponding to time of 0.002 ms to 0.004 ms. The first peak in the matched filtering output in the range is detected as the object surface. Duration of the echo is fixed for all the recording to 2.0 ms.

## 2.2 FORM FUNCTION CALCULATION

The Form Function expresses a pressure field scattered from a target in a range of frequencies [1]. It describes the way an object scatters a pulse, which makes the Form Function a good descriptor of an object itself.

In this work the Form Function is calculated from scattered and incident pulses with a knowledge about the medium where the signal is propagating. The frequency range of the function is limited by the bandwidth of the initial pulse.

The Form Function is computed based on [8], where the scattered echo is calculated from a Form Function and initial pulse. Reformulating the task, the Form Function can be found by Equation 1, where r is a distance between source and the target, $s(t)$ is a reflected pulse and $si(t)$ – initial  pulse.

$$f_\infty = \frac{\left|k_L^{(1)}\right| r^2}{e^{-2jk_L^{(1)}r}} \frac{FT[s(t)]}{FT[s_i(t)]}$$ (1)

The distance to the object $r$ is a parameter which has to be evaluated for each set of experimental data separately. The distance between the pulse source and the object is computed based on the matched filter output.

The distance $r$ is proportional to the distance between first peak and the peak detected as the scattering from the object of interest, $\Delta t$, Equation 2, where c is the speed of sound in the outer medium.

$$r = \Delta t \frac{c}{2}$$ (2)

In this work the Form Function is computed for a set of spherical objects. The objects divided into 2 classes: filled with air and filled with water. They were insonified with a 1 ms linear down-chirp pulses in a frequency range from 30 to 160 kHz.

Figure 2 illustrates an example of the Form Function calculated for the reflection from an aluminium sphere filled with water.

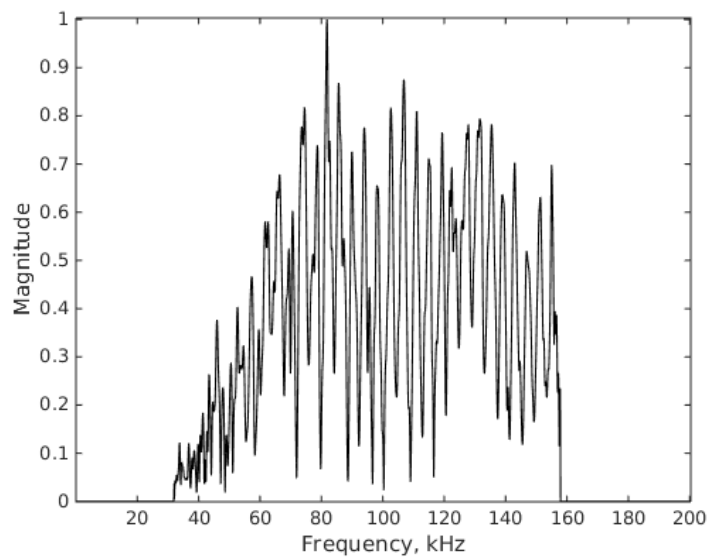

*Fig.2: Form Function of an aluminium sphere.*

The calculated Form Functions are used as descriptors for the classification.

## 2.3 CLASSIFICATION

The high performance of Neural Network classifiers was a motivation for the choice of a Multilayer Perceptron (MLP). The MLP has a 5 fully connected layers (FC) configuration, see Figure 3. The FC layers, except the output layer, are followed by Dropout layers with p=0.5 to prevent overfitting of the NN. ReLU activations are used for all the layers except the output layer, which uses sigmoid activation.

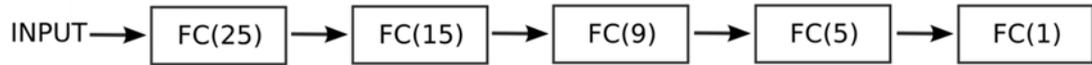

*Fig.3:* Multiplayer Perceptron architecture

The results of the MLP classifier are compared with the Support Vector Machine (SVM) classifier. SVM classifier is chosen as binary classifier with a Radial Basis Function kernel.

## 3. RESULTS

The data is classified into 2 classes with 430 examples per class. The classification is performed with 3-Fold cross-validation for both classifiers to eliminate effect of data on the classification accuracy and provide fair evaluation of the results.

Performance of the Form Function based classifiers is compared with performance of the echo in Time and Frequency Domains based classifiers. Results for both MLP and SVM classifiers are summarised in Table 1.

| Echo representation | Mean Accuracy ± Standard Deviation, % | |
|---|---|---|
| | MLP | SVM |
| Form Function | 98.60 ± 1.31 | 91.98 ± 1.78 |
| Frequency Domain | 92.43 ± 3.41 | 52.67 ± 5.28 |
| Time Domain | 97.21 ± 3.94 | 49.18 ± 1.96 |

*Table 1: Classification performance of MLP and SVM.*

The highest accuracy is achieved with Form Function based MLP classifier. The results show that MLP classifier outperforms SVM. The Form Function based classification results present higher accuracy for both classifiers. The performance of the SVM classifier demonstrates significant advance in the accuracy for the Form Function case, while MLP results have less critical difference. This difference in variation can be caused by the high performance of Neural Network classifiers in general.

## 4. CONCLUSION AND FUTURE WORK

This paper presents an approach for classification of spherical objects based on Form Function with a MLP Neural Network classifier. The objects are classified into 2 groups based on its filler material. The classifier is compared to the SVM classifier and Form Function descriptor is evaluated in contrast to the echo representation in Time and Frequency Domains. The Form Function based MLP classifier outperforms the other results showing 98% classification accuracy.

The results shows advantage of using the Form Function descriptor and illustrate that the Form Function is a stable feature vector which doesn't depend on the initial pulse and suitable for classification of object's materials.

For future work, we would like to interpret the Neural Network classifier. It would provide information about components of the Form Function which are participating in the classification and give a possibility to connect its structure to the physical parameters of the object.

## ACKNOWLEDGEMENTS

The authors would like to acknowledge financial support of the FP7-PEOPLE-2013-ITN project ROBOCADEMY (Ref 608096) funded by the European Commission.

## REFERENCES

[1] **M.J. Crocker**, Handbook of Acoustics, John Wiley and Sons, 1998.

[2] **James J. Faran**, Sound Scattering by Solid Cylinders and Spheres, the Journal of the Acoustic Society of America, vol. 23, pp. 405-417, 1951.

[3] **R. Hickling**, Analysis of echoes from a solid elastic sphere in water, the Journal of the Acoustical Society of America, vol. 34, pp. 1582–1592, 1962.

[4] **R. R. Goodman and R. Stern**, Reflection and transmission of sound by elastic spherical shells, the Journal of the Acoustical society of America, vol.34, 1962.

[5] **R. Doolittle and H. Uberall**, Sound scattering by elastic cylindrical shells, the Journal of the Acoustic Society of America, vol. 39, pp. 272–275, 1965.

[6] **P. A. Chinnery, V. F. Humphrey, and J. Zhang**, Low frequency acoustic scattering by a cube: Experimental measurements and theoretical predictions, The Journal of the Acoustical Society of America, vol. 101(5), 1997.

[7] **Y. Pailhas, C. Capus, K. Brown, and P. Moor**, Identifying content of low profile target in cluttered environment using biosonar," Underwater Acoustics conference, Rhodes, Greece, 2012.

[8] **Y. Pailhas, C. Capus, K. Brown, and Y. Petillot**, Design of artificial landmarks for underwater simultaneous localisation and mapping," IET Radar Sonar Navig., pp. 1–9, 2013.

[9] **G. C. Gaunaurd, D. Brill, H. Huangb, P. W. B. Moore, and H. C. Strifors**, Signal processing of the echo signatures returned by submerged shells insonified by dolphin clicks: active classification," The Journal of the Acoustical Society of America, vol.103(3), pp. 1547–1557, 1998,

[10] **A. Tesei, W. L. J. Fox, A. Maguer, and A. Lovik**, Target parameter estimation using resonance scattering analysis applied to air-filled, cylindrical shells in water, The Journal of the Acoustical Society of America, vol.108, 2000

[11] **W. Li, G. R. Liu, and V. K. Varadan**, Estimation of radius and thickness of a thin spherical shell in water using the midfrequency enhancement of a short tone burst response," The Journal of the Acoustical Society of America, vol.4, 118.